\documentclass[runningheads]{llncs}

\usepackage{eccv}

\usepackage{eccvabbrv}

\usepackage{amsmath}
\usepackage{booktabs} 
\usepackage{multirow} 
\usepackage{multicol} 
\usepackage{arydshln}
\usepackage{colortbl}
\usepackage{graphicx}
\usepackage{booktabs}

\usepackage[accsupp]{axessibility}  

\usepackage{hyperref}

\usepackage{orcidlink}

\begin{document}

\title{SegVG: Transferring Object Bounding Box to Segmentation for Visual Grounding}

\titlerunning{Transferring Object Bounding Box to Segmentation for Visual Grounding}

\author{Weitai Kang\inst{1}\orcidlink{0009-0007-6484-0665} \and
Gaowen Liu\inst{2} \and
Mubarak Shah\inst{3} \and
Yan Yan\inst{1}
}

\authorrunning{W.~Kang et al.}

\institute{Illinois Institute of Technology \and Cisco Research \and University of Central Florida}
\maketitle
\begin{abstract}
Different from Object Detection, Visual Grounding deals with detecting a bounding box for each text-image pair. This one box for each text-image data provides sparse supervision signals. Although previous works achieve impressive results, their passive utilization of annotation, i.e. the sole use of the box annotation as regression ground truth, results in a suboptimal performance. In this paper, we present {\bf SegVG}, a novel method transfers the box-level annotation as {\bf Seg}mentation signals to provide an additional pixel-level supervision for {\bf V}isual {\bf G}rounding. Specifically, we propose the Multi-layer Multi-task Encoder-Decoder as the target grounding stage, where we learn a regression query and multiple segmentation queries to ground the target by regression and segmentation of the box in each decoding layer, respectively. This approach allows us to iteratively exploit the annotation as signals for both box-level regression and pixel-level segmentation. Moreover, as the backbones are typically initialized by pretrained parameters learned from unimodal tasks and the queries for both regression and segmentation are static learnable embeddings, a domain discrepancy remains among these three types of features, which impairs subsequent target grounding. To mitigate this discrepancy, we introduce the Triple Alignment module, where the query, text, and vision tokens are triangularly updated to share the same space by triple attention mechanism. Extensive experiments on five widely used datasets validate our state-of-the-art (SOTA) performance.
Code is available at \url{https://github.com/WeitaiKang/SegVG}.
\end{abstract}    
\begin{figure}[t]
\centering
\includegraphics[width=0.8\textwidth]{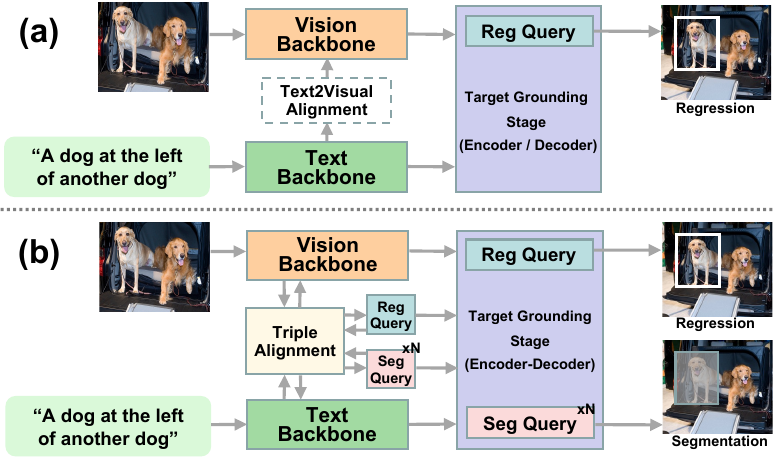}
\vspace{-8pt}
\caption{
The comparison of visual grounding frameworks. The block with a dashed border indicates that the module may not necessarily exist. ({\bf a}) Previous baseline method consists of two backbones and additional transformer layers for target grounding, where a regression query is supervised to regress the box. Current SOTA methods further employ a text-to-visual module to align the visual features with text features. ({\bf b}) Our method incorporates segmentation queries, which utilizes the box annotation at the pixel-level to segment the target. Additionally, we propose the Triple Alignment module to eliminate the domain discrepancy of the query, text, and vision features.}
\label{intro}
\vspace{-10pt}
\end{figure}

\section{Introduction}
Visual grounding~\cite{mao2016generation,plummer2015flickr30k,yu2016modeling, kazemzadeh2014referitgame,kang2024intent3d} aims to localize a target object within an image based on a free-form natural language text expression. It is particularly important for numerous downstream multimodal reasoning systems, such as visual question answering~\cite{gan2017vqs,wang2020general,shang2024llava} and image captioning~\cite{anderson2018bottom,chen2020say,you2016image}. Previous works can be broadly categorized into three distinct groups: two-stage methods~\cite{yang2019dynamic,yang2020graph,yu2018mattnet,chen2021ref}, one-stage methods~\cite{yang2019fast,yang2020improving}, and transformer-based ones~\cite{deng2021transvg,ye2022shifting,yang2022improving,qu2022siri,kang2024visual,kang2024actress}. Both two-stage and one-stage approaches use convolutional neural networks for candidate proposals and the selection of the best-matching candidate. Nonetheless, these approaches rely on intricate modules that employ manually-crafted techniques for performing language inference and multi-modal integration.

Inspired by the success of the transformer \cite{devlin2018bert, dosovitskiy2020image}, TransVG~\cite{deng2021transvg} proposes a transformer-based pipeline. As shown in Fig.~\ref{intro}.(a), this pipeline extracts vision and text features via DETR~\cite{carion2020end} and BERT~\cite{devlin2018bert}, respectively. To ground the target, they use the transformer encoder to fuse multimodal features along with a learnable regression query and decode the query through an MLP. To enhance the final target grounding stage, subsequent studies continue with some text-to-visual modules in the early stage to modulate the vision features to align with the text features. For example, QRNet~\cite{ye2022shifting} proposes a query-modulated method for extracting language-aware vision features within the vision backbone. VLTVG~\cite{yang2022improving} introduces a verification map to activate the vision features to align with the text features before multimodal fusion.

Despite their advancements, the suboptimal annotation utilization, i.e., only using the box annotation as a regression annotation, limits their performance. As discussed in \cite{sun2023refteacher}, Visual Grounding presents unique challenges compared to Object Detection due to its sparse supervision signals. Specifically, it provides only one box label for each text-image pair, while necessitatng detection within a multimodal setting. Therefore, it is essential to fully exploit the box annotation, by treating it as a segmentation mask (pixels within the bounding box are assigned a value of 1, while pixels outside the bounding box are assigned as 0).

In this paper, we introduce SegVG (see  Fig.~\ref{intro}.(b)), a novel method that leverages the pixel-level details within the box annotation as segmentation signals to offer additional fine-grained supervision for Visual Grounding. Specifically, we propose the Multi-layer Multi-task Encoder-Decoder as the target grounding stage, where we learn a regression query and multiple segmentation queries to ground the target by regression and segmentation of the box in each decoding layer, respectively. The confidence score derived from the segmentation can further serve as a Focal Loss~\cite{lin2017focal} scaling factor to adaptively emphasize the other losses of challenging training samples. This approach allows us to iteratively exploit the annotation as signals for both box-level regression and pixel-level segmentation. Furthermore, the initial parameters for model backbones, typically derived from pretrained unimodal tasks, along with data-agnostic static embeddings used as queries for decoding, result in a domain discrepancy among different sources of feature, affecting the effectiveness of target grounding. To tackle this problem, we present the Triple Alignment module, where we harmonize the domain of query, text, and vision features by implementing a triangular update process through a triple attention mechanism. As a result, we ensure that all features adapt and integrate within the same multimodal space, thereby enhancing subsequent target grounding. Our contributions are as follows:
\begin{itemize}
  \item We propose the Multi-layer Multi-task Encoder-Decoder to maximize the utilization of the box annotation, which introduces an additional segmentation format for pixel-level supervision in Visual Grounding.
  \item To eliminate the domain discrepancy among the query, text, and vision, we introduce the Triple Alignment to update these three types of features into a sharing domain, which facilitates the subsequent target grounding.
  \item We conduct extensive experiments on five widely used datasets to show the performance superiority of our proposed methods compared with previous state-of-the-art methods and further investigate the reliability benefits derived from the segmentation output in real applications.
  \item We will release source code and checkpoints upon acceptance of the paper for future research development.
  \vspace{-18pt}
\end{itemize}

\begin{figure}[t]
\centering
\includegraphics[width=0.98\textwidth]{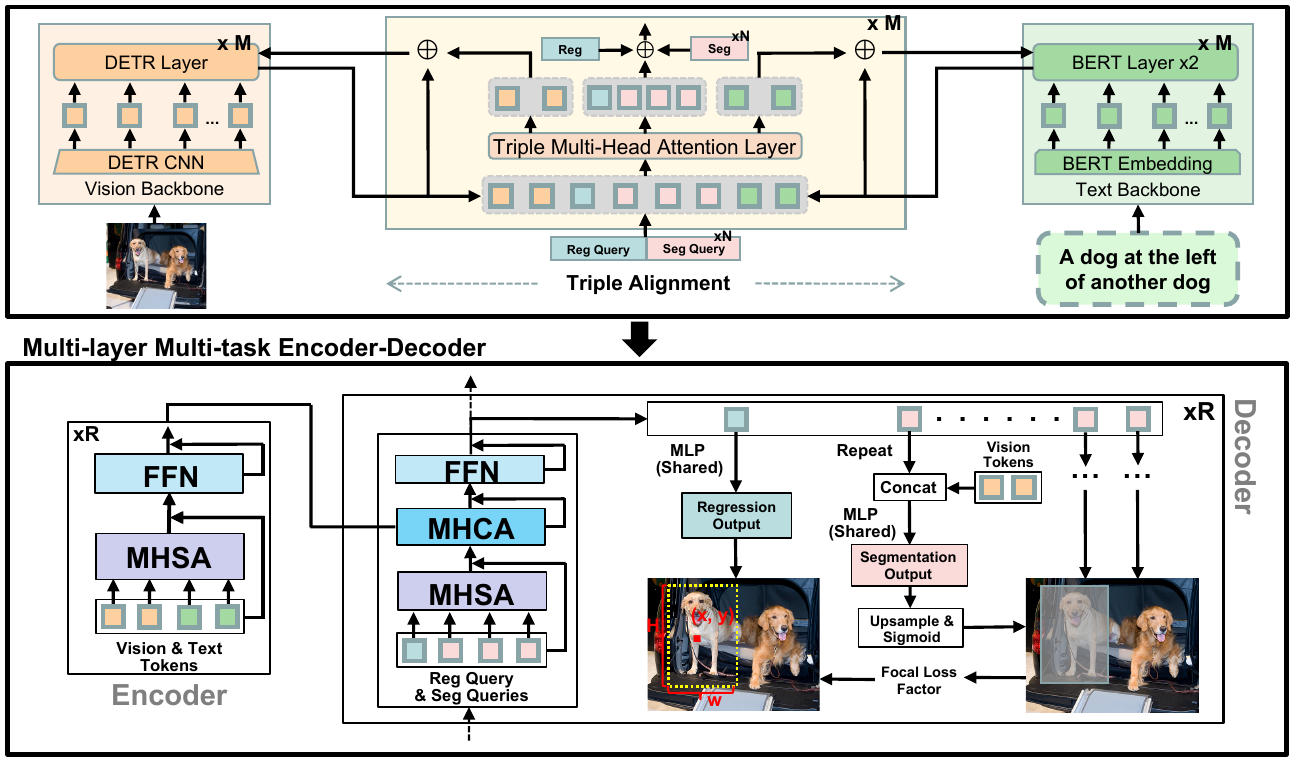}
\vspace{-8pt}
\caption{
SegVG: {\bf The upper figure} includes the vision and text backbone. Our proposed Triple Alignment module is iteratively inserted into intermediate layers to eliminate domain discrepancy. {\bf The lower figure} shows our Multi-layer Multi-task Encoder-Decoder, which adopts a transformer encoder-decoder to update multimodal features and ground the target. In this architecture, we make the best of the box annotation as a segmentation ground truth and integrate an additional segmentation task into Visual Grounding. Additionally, the segmentation output serves as a Focal Loss factor, allowing adaptive emphasis on challenging cases for the regression loss. M = 6, R=6.
}
\label{method}
\vspace{-10pt}
\end{figure}

\section{Related Work}
Visual grounding methods can be roughly classified into three pipelines: two-stage methods, one-stage methods, and transformer-based methods.

\noindent \textbf{Two-stage methods}
Two-stage approaches~\cite{yu2018mattnet, chen2021ref} treat visual grounding as first generating candidate object proposals and then finding the best match to the text. In the first stage, an off-the-shelf detector processes the image and proposes regions that may contain the target. In the second stage, a ranking network calculates the similarity between candidate regions and processed text features, selecting the region with the highest similarity score as the final result. Training losses include binary classification loss~\cite{plummer2018conditional} or maximum-margin ranking loss~\cite{yu2018mattnet}. To better understand the text and cross-modality matching, MattNet~\cite{yu2018mattnet} focuses on decomposing the text into subject, location, and relationship components. \cite{chen2021ref} introduces an expression-aware score for improved candidate region ranking. 

\noindent \textbf{One-stage methods}
One-stage approaches~\cite{yang2019fast, yang2020improving} directly concatenate vision and text features in the channel dimension and rank confidence values for candidate regions proposed based on the concatenated multimodal features. For example, FAOA~\cite{yang2019fast} predicts bounding boxes using a YOLOv3 detector~\cite{redmon2018yolov3} on the concatenated features. ReSC~\cite{yang2020improving} further improves the ability to ground complex queries by introducing a recursive sub-query construction module. 

\noindent \textbf{Transformer-based methods}
Transformer-based approach is first introduced by TransVG~\cite{deng2021transvg}. Unlike previous methods, TransVG concatenates the regression query (a learnable embedding), vision tokens, and text tokens and uses transformer encoders~\cite{vaswani2017attention} to perform cross-modal fusion and target grounding. The query is then processed through an MLP to decode the box. Benefiting from the flexible structure of transformer modules in processing multimodal features, recent works continue adopting this pipeline and propose novelties regarding feature extraction. 
VLTVG~\cite{yang2022improving} develops a visual-linguistic verification module before the target grounding stage to modulate the vision features with the relationship between vision and text features. QRNet~\cite{ye2022shifting} proposes a Query-modulated Refinement Network to early fuse visual and text features, mitigating the gap between features from the unimodal vision backbone and those needed for multi-modal reasoning.

\noindent \textbf{Multi-task Visual Grounding}
Multi-task learning is extensively utilized in object detection and segmentation \cite{carion2020end, he2017mask}, often capitalizing on a shared backbone and task-specific heads. Expanding upon this idea, several studies \cite{luo2020multi, li2021referring, su2023language} have proposed solutions to the Multi-task Visual Grounding problem. In this problem, they jointly tackle Referring Expression Comprehension (REC, also known as Visual Grounding) and Referring Expression Segmentation (RES), requiring both box annotations and segmentation annotations. It is important to note that, unlike those approaches, even though we incorporate segmentation losses in our method, \textbf{we do not need segmentation annotations} but only box annotations, focusing specifically on the Visual Grounding task.
\section{Methodology}
In this section, we present the components of our SegVG in the order of the data flow: starting with the backbones, followed by our proposed Triple Alignment, and finally our Multi-layer Multi-task Encoder-Decoder.

\subsection{Backbones}
Similar to previous works~\cite{deng2021transvg, yang2022improving}, as shown in Fig.~\ref{method} (upper), our vision backbone consists of ResNet and transformer encoder from DETR~\cite{carion2020end}, with parameters pretrained on Object Detection task using the MSCOCO dataset~\cite{lin2014microsoft}, excluding the validation and test sets of Visual Grounding dataset. The text backbone is the base model of BERT~\cite{devlin2018bert}.

\paragraph{Vision Backbone} Given an input image ${\bf I_0}$ ($  \mathbb{R}^{3\times {H_0}\times {W_0}}$), we employ the DETR's ResNet to generate a 2D feature map ${\bf I} \in \mathbb{R}^{C\times {H}\times {W}}$ $(C = 2048, H = \frac{H_0}{32}, W = \frac{W_0}{32})$. A 1x1 convolutional layer is then used to reduce the channel dimension of $\bf I$ to $C_v = 256$, resulting in ${\bf I^{'}}$. We further flatten ${\bf I^{'}}$ into ${\bf Z}_v \in \mathbb{R}^{C_v\times{N_v}}$ $({N_v} = H \times W)$. Position embedding is then added to ${\bf Z}_v$ to preserve sensitivity to the original 2D spatial locations. ${\bf Z}_v$ is then iteratively processed through DETR's encoder layer (total 6 transformer layers) and the Triple Alignment to obtain the output ${\bf Z}_v$.

\paragraph{Text Backbone} Given a text, we initially utilize the BERT's embedding layer to convert it into $N_t$ language tokens with $C_t$ channel dimension. In alignment with \cite{deng2021transvg}, we prepend a [CLS] token and append a [SEP] token to the beginning and end positions of the tokenized language, respectively. Following this, we iteratively input the language tokens into BERT's layers (total 12 transformer layers) and the Triple Alignment, generating language embedding ${\bf Z}_t$.

\subsection{Triple Alignment}
Given that the text and vision backbones are pretrained from unimodal tasks and the queries are data-agnostic, the subsequent target grounding stage faces the challenge of aligning these three types of features into the same space before performing multimodal fusion for target grounding. Additionally, considering that the backbones usually contribute the majority of the overall parameters, solely using them for extracting unimodal features without incorporating multimodal alignment is sub-optimal. Therefore, an optimal solution is to address the domain discrepancy before moving on to the subsequent target grounding stage.

As shown in Fig.~\ref{method} (upper), our proposed Triple Alignment module utilizes an attention mechanism to perform triangular feature sampling, aiming to ensure domain consistency among the query, text, and vision features. The queries, ${\bf Z}_{o}$, are first initialized by N learnable embeddings, where one embedding is for the regression query and the rest of the embeddings are for multiple segmentation queries. The data flow is:
\begin{equation}
    {\bf Z}_{v}^{i+1} = {\rm DETRLayer_\textit{i}}({\bf Z}_{v}^i),  i \in \{0, 1, .., L-1\}
    \label{eq1}
    \vspace{-14pt}
\end{equation}
\begin{equation}
    {\bf Z}_{t}^{i+1} = {\rm BERTLayer_\textit{2i+1}}({\rm BERTLayer_\textit{2i}}({\bf Z}_{t}^i)),
    \label{eq2}
    \vspace{-10pt}
\end{equation}
\begin{equation}
    [{\bf Z}_{o}^{'}, {\bf Z}_{t}^{'}, {\bf Z}_{v}^{'}] = {\rm Tri\textsc{-}MHA}({\bf Z}_{o}, {\bf Z}_{t}^{i+1}, {\bf Z}_{v}^{i+1}),
    \label{eq3}
    \vspace{-10pt}
\end{equation}
\begin{equation}
    {\bf Z}_{o} = {\bf Z}_{o} + {\bf Z}_{o}^{'},
    {\bf Z}_{t}^{i+1} = {\bf Z}_{t}^{i+1} + {\bf Z}_{t}^{'},
    {\bf Z}_{v}^{i+1} = {\bf Z}_{v}^{i+1} + {\bf Z}_{v}^{'},
\label{eq4}
\end{equation}
where L is the number of layers, BERTLayer is the layer of BERT and DETRLayer is the layer of DETR's encoder. The vision and text features are first encoded by Eq.~\ref{eq1} and Eq.~\ref{eq2}. Subsequently, the three types of tokens (query, text, and vision) are updated by our Triple Multi-Head Attention Layer (Tri-MHA) using Eq.~\ref{eq3}. The output tokens are merged back to their original branches respectively by Eq.~\ref{eq4}. Within each head of the Triple Multi-Head Attention Layer (Tri-MHA), each type of the features simultaneously computes its updated representation by attending to both the others and itself:
\begin{equation}
\begin{aligned}
S=[{\bf Z}_{o}W^{(o, S)},&{\bf Z}_{t}W^{(t, S)},{\bf Z}_{v}W^{(v, S)}],S\in\{Q,K,V\} \\
[{\bf Z}_{o}, {\bf Z}_{t}, &{\bf Z}_{v}] = {\rm SoftMax}(QK^{T} / {\sqrt d_{k}})V, \\
{\bf Z}_{e}^{'}& =~ {\bf Z}_{e}W^{e}, e \in \{o, t, v\}, \\
\end{aligned}
\label{eq5}
\end{equation}
where $\{W^{(e, S)}, W^{e}: e \in \{o, t, v\}, S \in \{Q, K, V\}\}$ are trainable parameter. As a result, each of the output features is triangular sampling from all of the three types of features, which alleviates the domain discrepancy. 

\subsection{Multi-layer Multi-task Encoder-Decoder}
The Multi-layer Multi-task Encoder-Decoder serves as the target grounding stage, where we use a transformer encoder-decoder for cross-modal fusion and target grounding to perform a box regression task and a box segmentation task. 

\paragraph{Encoder} 
As shown in Fig.~\ref{method} (lower left), given the aligned output text and vision features from the backbones, the encoder fuses the two modalities into the multimodal features by a stack of transformer layers. In each layer, the concatenated text and vision tokens go through the Multi-Head Self-Attention layer (MHSA) and the Feed Forward Network (FFN) with the residual connection.

\paragraph{Decoder} 
In each decoder layer, we aim to fully exploit the box annotation. We propose the {\bf bbox2seg} paradigm to transform the box annotation into a segmentation mask, which classifies all pixels within the box as foreground (with a value of one) and those outside the box as background (with a value of zero). As shown in Fig.~\ref{method} (lower right), one regression query aims to regress the box, while the remaining segmentation queries aim to segment the box. Different segmentation queries are endowed with different learnable positional embeddings to enhance the robustness of each decoder layer, since the decoder layer, when confronted with various queries, is required to segment the same box. Following that, the queries pass through the Multi-Head Self-Attention layer to exchange information about the same target, prompting each other to better locate the target. Subsequently, the queries undergo the Multi-Head Cross-Attention layer and the Feed Forward Network, where multimodal features serve as the Key and Value to ground the target. Finally, a shared MLP across all decoder layers decodes the regression query into the box result, supervised by L1 loss and Giou loss\cite{rezatofighi2019generalized}. Each segmentation query is repeated $N_v$ times and concatenated with visual tokens along the channel dimension. Another shared MLP decodes the concatenated feature into the segmentation mask, supervised by Focal loss~\cite{lin2017focal} and Dice loss~\cite{milletari2016v}. It is noteworthy that \textbf{our segmentation paradigm shares the same semantic foundation as the regression paradigm, i.e., to distinguish bounding box}, rather than instance segmentation. Therefore, incorporating non-object pixels in the segmented foreground does not introduce ambiguity to the model. We provide qualitative results~\ref{qualitative_res} to demonstrate this feature. To alleviate multi-task optimization challenges, we freeze the backbones for the initial k epochs to stabilize the training process.

\paragraph{Confidence score} 
Since both the regression output and segmentation output share the same aim, we can additionally obtain the confidence score for the foreground by averaging values inside the ground truth box of the segmentation output to reflect the confidence of the regression output. In the training process, we can transform this confidence score as the Focal loss factor~\cite{lin2017focal} to adaptively emphasize the other losses of challenging training samples. The final loss function of each decoder layer is formulated as follows:
\begin{equation}
\begin{aligned}
L = \lambda_{1} c_{focal} L_1 + \lambda_{giou} c_{focal} L_{giou} + \\
\lambda_{dice} c_{focal} L_{dice} + \lambda_{focal} L_{focal},
\end{aligned}
\label{loss}
\end{equation}
where $\lambda_{1}$, $\lambda_{giou}$, $\lambda_{focal}$ and $\lambda_{dice}$ are hyperparameters. $L_1$ is the L1 loss. $L_{giou}$ is the GIoU loss \cite{rezatofighi2019generalized}. $L_{focal}$ is the Focal loss \cite{lin2017focal}. $L_{dice}$ is the Dice loss \cite{milletari2016v}. $c_{focal}$ is the above Focal loss factor averaged across all segmentation outputs.

In the real-world application perspective, the visual grounding task can be viewed as open-vocabulary object detection~\cite{zareian2021open}, where target objects lack predetermined categories. Therefore, previous transformer-based methods directly regress the box without confidence scores, since there is no candidate proposal or selection stage in transformer-based pipeline. However, confidence scores are valuable for enhancing the control or reliability of predictions by filtering out low-confidence predictions. This feature could benefit the future integration of visual grounding models into downstream multimodal reasoning systems or real-world applications. To meet the requirements of this feature, our approach incorporates a confidence score derived from the segmentation output during inference. Specifically, we calculate the model's confidence by averaging values greater than or equal to 0.35 (adopted from~\cite{wang2022cris}) in the segmentation output of one segmentation query. 
Analyses~\ref{conf_analyse_sec} in the Experiment section demonstrate the faithfulness and benefits of incorporating this additional confidence score.
\section{Experiments}
\vspace{-5pt}

\begin{table}[t]
\centering
\caption{Comparisons with state-of-the-art methods on widely used datasets. We highlight the best and second best performance in {\bf\textcolor{red}{red}} and {\textcolor{blue}{blue}}, and bold our model.}
\vspace{-10pt}
\resizebox{1\linewidth}{!}{%
\setlength{\tabcolsep}{0.5mm}
\label{table1}
\begin{tabular}{c|c|ccc|ccc|ccc|c}
\hline
\multirow{2}{*}{Models} & \multirow{2}{*}{Backbone} & \multicolumn{3}{c|}{RefCOCO} & \multicolumn{3}{c|}{RefCOCO+} & \multicolumn{3}{c|}{RefCOCOg} & ReferItGame \\
 & &\textit{val} &\textit{testA} &\textit{testB} &\textit{val} &\textit{testA} &\textit{testB} &\textit{val-g} &\textit{val-u} &\textit{test-u} & \textit{test} \\
\hline
\textit{Two-stage:} &  & & & & & &  & & & & \\
CMN~\cite{hu2017modeling} & VGG16 &- &71.03 &65.77 &- &54.32 &47.76 &57.47 &- &- & 28.33 \\
VC~\cite{zhang2018grounding} & VGG16 &- &73.33 &67.44 &- &58.40 &53.18 &62.30 &- &-& 31.13 \\
ParalAttn~\cite{zhuang2018parallel} & VGG16 &- &75.31 &65.52 &- &61.34 &50.86 &58.03 &- &- &- \\
MAttNet~\cite{yu2018mattnet} & ResNet-101 &76.65 &81.14 &69.99 &65.33 &71.62 &56.02 &- &66.58 &67.27 &29.04 \\
Similarity Net~\cite{wang2018learning} & ResNet-101 &-&-&-&-&-&-&-&-&-&34.54\\
CITE~\cite{plummer2018conditional} & ResNet-101 &-&-&-&-&-&-&-&-&-&35.07 \\
DDPN~\cite{yu2018rethinking} & ResNet-101  &-&-&-&-&-&-&-&-&-&63.00 \\
LGRANs~\cite{wang2019neighbourhood} & VGG16 &- &76.60 &66.40 &- &64.00 & 53.40 &61.78 &- &-&-\\
DGA~\cite{yang2019dynamic} & VGG16 &- &78.42 &65.53 &- &69.07 &51.99 &- &- &63.28&-\\
RvG-Tree~\cite{plummer2018conditional} & ResNet-101 &75.06 &78.61 &69.85 &63.51 &67.45 &56.66 &- &66.95 &66.51&-\\
NMTree~\cite{liu2019learning} & ResNet-101 &76.41 &81.21 &70.09 &66.46 &72.02 &57.52 &64.62 &65.87 &66.44&-\\
Ref-NMS~\cite{chen2021ref} & ResNet-101 &80.70 &84.00 &76.04 &68.25 &73.68 &59.42 &- &70.55 &70.62&-\\
\hline
\textit{One-stage:} &  & & & & & &  & & & & \\
SSG~\cite{chen2018real} & DarkNet-53 &- &76.51 &67.50 &- &62.14 &49.27& 47.47& 58.80& - &54.24\\
ZSGNet~\cite{sadhu2019zero} & ResNet-50&-&-&-&-&-&-&-&-&-& 58.63 \\
FAOA~\cite{yang2019fast} & DarkNet-53& 72.54& 74.35& 68.50 &56.81& 60.23& 49.60 &56.12 &61.33 &60.36&60.67\\
RCCF~\cite{liao2020real} & DLA-34& - &81.06& 71.85 &- &70.35 &56.32& - &- &65.73&63.79  \\
ReSC-Large~\cite{yang2020improving}&  DarkNet-53& 77.63& 80.45& 72.30& 63.59& 68.36& 56.81& 63.12& 67.30& 67.20&64.60 \\
LBYL-Net~\cite{huang2021look} & DarkNet-53 &79.67 &82.91 &74.15 &68.64 &73.38 &59.49 &62.70 &- &-&67.47 \\
\hline
\textit{Transformer-based:} &  & & & & & &  & & & & \\
TransVG~\cite{deng2021transvg} & ResNet-101 &81.02 &82.72 &78.35 &64.82 &70.70 &56.94 &67.02 &68.67 &67.73&70.73 \\
QRNet~\cite{ye2022shifting} & Swin-S 
&84.01 &85.85 &\textcolor{blue}{82.34} 
&72.94 &76.17 &63.81 
&71.89 &73.03 &72.52 
&\textcolor{blue}{74.61} \\
VLTVG~\cite{yang2022improving} &ResNet-101
& \textcolor{blue}{84.77}& \textcolor{blue}{87.24} &80.49 
&\textcolor{blue}{74.19} &\textcolor{blue}{78.93} &\textcolor{blue}{65.17} 
&\textcolor{blue}{72.98} &\textcolor{blue}{76.04} &\textcolor{blue}{74.18}
& 71.98\\
SegVG (ours) & ResNet-101 
&\bf\textcolor{red}{86.84}  &\bf\textcolor{red}{89.46}  &\bf\textcolor{red}{83.07}
&\bf\textcolor{red}{77.18}  &\bf\textcolor{red}{82.63}  &\bf\textcolor{red}{67.59} 
&\bf\textcolor{red}{76.01}  &\bf\textcolor{red}{78.35}  &\bf\textcolor{red}{77.42} 
&\bf\textcolor{red}{75.59} \\
\hline
\end{tabular}
}
\vspace{-14pt}
\end{table}

\subsection{Metric and Datasets}
\vspace{-4pt}
\paragraph{Metric} A predicted bounding box is considered accurate if its Intersection over Union (IoU) with the ground-truth bounding box exceeds 0.5. In accordance with the established practices in preceding studies~\cite{deng2021transvg,yang2022improving}, we employ top-1 accuracy (measured in percentage) as the primary metric to assess our method. 

\paragraph{Datasets} 
\sloppy
There are five standard benchmarks: RefCOCO \cite{yu2016modeling}, RefCOCO+ \cite{yu2016modeling}, RefCOCOg-g \cite{mao2016generation}, RefCOCOg-umd \cite{mao2016generation}, and ReferItGame \cite{kazemzadeh2014referitgame}. Four of them (RefCOCO, RefCOCO+, and RefCOCOg-(g/umd)) are all derived from MSCOCO~\cite{lin2014microsoft}. RefCOCO consists of 19,994 images and 142,210 referring texts, which is divided into four subsets: a training set with 120,624 texts, a validation set with 10,834 texts, and two test sets (testA and testB) containing 5,657 and 5,095 texts, respectively. RefCOCO+ includes 19,992 images and 141,564 referring texts, which is partitioned into four subsets: a training set with 120,191 texts, a validation set with 10,758 texts, and two test sets (testA and testB) containing 5,726 and 4,889 texts, respectively. RefCOCOg contains 25,799 images and 95,010 longer texts. Two widely accepted splitting methods—RefCOCOg-g~\cite{mao2016generation} and RefCOCOg-umd~\cite{nagaraja2016modeling}—are employed for this dataset, and we perform experiments using both RefCOCOg-g (val-g) and RefCOCOg-umd (val-u and test-u) splitting conventions. The ReferItGame dataset, featuring 20,000 images from SAIAPR-12~\cite{escalante2010segmented}, is divided into three segments: a training set with 54,127 texts, a validation set with 5,842 texts, and a testing set comprising 60,103 texts.

\subsection{Implementation}

\paragraph{Input Configuration} Our approach uses an input image size of 640 x 640 and sets the maximum expression length at 40. When resizing images, we preserve the original aspect ratio. The longer edge is resized to 640, and the shorter edge is padded to 640 with the value of zero. Texts exceeding 38 tokens are truncated, reserving a start position and an end position of the characters for the [CLS] and [SEP] token, respectively. If the text is shorter, empty tokens are added after the [SEP] token to reach an input length of 40. Paddings for the input image are not tracked by masks, while empty tokens of text employ masks. 

\paragraph{Training Procedure} We employ the AdamW optimizer. The initial learning rate of 1e-5 is assigned to the visual and language backbone, and 1e-4 to the rest parameters. Weight decay is set at 1e-4. The visual backbone is initialized with the DETR model's backbone and encoder, while the language branch uses the basic BERT model. For the final results, our model is trained for 90 epochs, with the learning rate decreasing by a factor of 10 after 60 epochs. The k hyperparameter in Multi-layer Multi-task Encoder-Decoder is set to 10. We use a batch size of 64. For the ablation studies presented in Table~\ref{ablation}, the models are trained for 60 epochs with k equal to 20, and the learning rate drops after 40 epochs. We set $\lambda_{1}$ = 5, $\lambda_{giou}$ = 2, $\lambda_{focal}$ = 1 and $\lambda_{dice}$ = 1. We adhere to previous practices~\cite{deng2021transvg, yang2022improving} for data augmentation during training.

\subsection{Quantitative Results}

We report the performance of our SegVG on all benchmark datasets. As presented in Table~\ref{table1}, our SegVG model demonstrates superiority across all of the datasets. This indicates the effectiveness and generalizability of our approach. It is worth noting that RefCOCO+ and RefCOCOg are relatively more challenging datasets, as RefCOCO+ does not include location terms in its language expressions, and RefCOCOg has longer language expressions compared to other datasets. Despite these challenges, our model exhibits significant improvements on these two difficult datasets. 
Specifically, on RefCOCO+, our model outperforms the previous SOTA models with +2.99\%, +3.7\%, and +2.42\% on the val, testA, and testB subsets, respectively. On RefCOCOg, our model also surpasses the previous SOTA models with +3.03\%, +2.31\%, and +3.24\% on the val-g, val-u, and test-u subsets, respectively. These results suggest that under the reinforcement of Triple Alignment and Multi-layer Multi-task Encoder-Decoder, the query, text, and vision tokens are triangularly updated to share the same space, and the model fully exploits the bounding box as fine-grained pixel-level supervision for comprehensive end-to-end learning.

\begin{table}[t] 
\centering
\caption{Compare transformer-based models on Parameter count and GFLOPS.}
\vspace{-10pt}
\resizebox{0.9\linewidth}{!}{%
\setlength{\tabcolsep}{4mm}
\begin{tabular}{c|c|c|c}
\hline
& Backbone & Parameter count (M) & GFLOPS (G) \\
\hline
TransVG~\cite{deng2021transvg} & ResNet101 & 141.55&  72.61 \\
\hline
VLTVG~\cite{yang2022improving} & ResNet101 & 141.61 & 69.87 \\
\hline
QRNet~\cite{ye2022shifting} & Swin-S & 247.06 & 80.12 \\
\hline
\rowcolor{gray!30} SegVG & ResNet101 & 155.28 & 73.48 \\
\hline
\end{tabular}}
\label{table2}
\end{table}

We also conduct a comparison of the number of parameters and GFLOPS across transformer-based models to evaluate computational costs. As depicted in Table \ref{table2}, the computational cost of SegVG falls within a reasonable range.

\subsection{Ablation Study}

In this section, we aim to validate the efficacy of each proposed module. We conduct ablation studies on the RefCOCOg-umd test dataset. Specifically, we start by evaluating a basic structure, i.e., the backbones with encoder-decoder structure. After that, we systematically incorporate the Triple Alignment module into the backbones and introduce the Multi-layer Multi-task supervision into the decoder through the controlled variable approach. Meanwhile, we conduct additional ablation experiments on specific details, including assessing the efficacy of incorporating the encoder, introducing Query in the Triple Alignment, and introducing the Focal loss from the segmentation output to the other losses.

\begin{table}[t]
  \centering
  \caption{Ablation study on RefCOCOg-umd test set. Encoder and Decoder are the encoder and decoder of Multi-layer Multi-task Encoder-Decoder, respectively. MMDecoder represents Multi-layer and Multi-task supervision in the Decoder. N represents the number of segmentation queries. Triple indicates the Triple Alignment. Excluding Query in Triple (in the (g)) represents the Triple Alignment is changed to bidirectional alignment with concatenated text-vision tokens input.
  }
  \vspace{-10pt}
  \resizebox{0.86\linewidth}{!}{
  \setlength{\tabcolsep}{1mm}
  \begin{tabular}{clcc}
    \toprule
    Id & Model & Acc(\%) \\
    \midrule
    (a) & Backbones + Encoder + Decoder & 66.97 \\
    (b) & Backbones + Triple + Encoder + Decoder & 75.75 \\
    (c) & Backbones + Encoder + MMDecoder (N=1) & 72.61 \\
    (d) & Backbones + Encoder + MMDecoder (N=5) & 76.21 \\
    (e) & Backbones + Triple + Encoder + MMDecoder (N=5) & \textcolor{red}{77.29} \\
    \midrule
    \midrule
    (f) & (e) w/o Encoder & 76.65\textcolor{blue}{(-0.64)} \\
    (g) & (e) w/o Query in Triple& 76.37\textcolor{blue}{(-0.92)} \\
    (h) & (e) w/o Focal loss & 76.83\textcolor{blue}{(-0.46)} \\
    \bottomrule
  \end{tabular}}
  \label{ablation}
  \vspace{-12pt}
\end{table}

\begin{table}[t]
  \centering
  \caption{Ablation study on RefCOCOg-umd test set regarding the number of segmentation queries.
  $\Delta$Hour and $\Delta$Acc are the additional time cost and accuracy improvement compared to ID(i) experiment, respectively.
  }
  \vspace{-10pt}
  \setlength{\tabcolsep}{2mm}
  \begin{tabular}{c|c|c|c|c}
  \specialrule{.1pt}{.1pt}{.1pt}
  ID & num\_query & Acc & Time Cost & $\Delta$Hour / $\Delta$Acc (v.s. ID(i)) \\
  \specialrule{.1pt}{.1pt}{.1pt}
  \rowcolor{gray!10} i   & N = 1 & 72.61 & 16.32h & - \\
  \rowcolor{gray!20} ii  & N = 3 & 73.02 & 18.97h & 6.46 \\
  \rowcolor{gray!30} iii & N = 5 & 76.21 & 20.10h & 1.05 \\
  \rowcolor{gray!20} iv  & N = 7 & 74.38 & 22.24h & 3.34 \\
  \rowcolor{gray!10} v  & N = 9 & 73.91 & 24.07h & 5.96 \\
  \specialrule{.1pt}{.1pt}{.1pt}
  \end{tabular}
  \label{ablation2}
  \vspace{-12pt}
\end{table}

As shown in Table \ref{ablation}, we can draw the following conclusions when comparing the experimental results under controlled variables:
\textbf{1) [(a)~v.s.~(b)]}: Incorporating the Triple Alignment can effectively eliminate the domain discrepancy among the query, text, and vision features, thereby facilitating subsequent target grounding. 
\textbf{2) [(a)~v.s.~(c)]}: Introducing Multi-layer Multi-task supervision can iteratively make the best of the annotations in the target grounding stage, thereby enhancing the learning of query representations.
\textbf{3) [(c)~v.s.~(d)]}: Increasing the number of segmentation queries can further improve the robustness of the decoder when provided with different queries and required to segment the same box.
\textbf{4) [(a), (b), (d), and (e)]}: Combining the Triple Alignment and Multi-layer Multi-task Encoder-Decoder can effectively enhance the overall performance, achieving the optimal result.
\textbf{5) [(e)~v.s.~(f)]}: Even if we include the Triple Alignment supporting multimodal communication, 
it remains necessary for the subsequent encoder to update the unimodal features generated by the backbones into multimodal features.
\textbf{6) [(e)~v.s.~(g)]}: It is necessary to involve queries in Triple Alignment to transfer the data-agnostic embedding into data-related queries. Otherwise, using only Bidirectional Alignment (BA) for text and vision tokens, similar to approaches like \textit{Deep Fusion} in GLIP~\cite{li2022grounded} and Grounding DINO~\cite{liu2023grounding}, and \textit{WPA} in CoupAlign~\cite{zhang2022coupalign}, causes a noticeable decline (-0.92\%).
\textbf{7) [(d), (e) and (g)]}: The slight gain of (g)(76.37\%) over (d)(76.21\%) stems from our MMDecoder's pixel-level signals, which already boosts BA in the Encoder. Thus, extra BA effort in the backbone is marginal, failing to solve the unaligned query issue. Instead,Tri-Align (e)(77.29\%) can solve this issue, showing its novelty. Notably, (d)(76.21\%), with a basic encoder-decoder, already achieves SOTA performance, emphasizing our bbox2seg paradigm's simplicity and effectiveness. \textbf{8) [(e)~v.s.~(h)]}: The segmentation output can further derive the confidence score of the model's prediction, which is transformed into the Focal loss factor to adaptively scale the other losses to focus more attention on challenging cases.

We further conduct a more detailed abaltion study about the improvement and corresponding cost of adding more segmentation queries of Tab.3{\bf (c)-(d)}. As shown in Table \ref{ablation2}, the best performance is observed with five segmentation queries. Adding more than that increases the burden from pixel-level constraint without benefits, also increasing computational cost per accuracy improvement.

\begin{table}[t] 
  \centering
  \caption{
Attention values from query to text-referred visual region in Triple Alignment.
  }
  \vspace{-10pt}
  \resizebox{1\linewidth}{!}{%
  \setlength{\tabcolsep}{1.2mm}
  \begin{tabular}{c|ccc|ccc|ccc|c}
  \hline
  \multirow{2}{*}{Layer} & \multicolumn{3}{c|}{RefCOCO} & \multicolumn{3}{c|}{RefCOCO+} & \multicolumn{3}{c|}{RefCOCOg} & ReferItGame \\
   & \textit{val} &\textit{testA} &\textit{testB} &\textit{val} &\textit{testA} &\textit{testB} &\textit{val-g} &\textit{val-u} &\textit{test-u} & \textit{test} \\
  \hline
  layer2 & 17\% & 18\% & 16\% & 19\% & 21\% & 18\% & 20\% & 18\% & 17\% & 25\% \\
  \rowcolor{gray!10} layer4 & 34\% & 36\% & 33\% & 29\% & 31\% & 28\% & 27\% & 33\% & 33\% & 30\% \\
  \rowcolor{gray!30} layer6 & 61\% & 63\% & 60\% & 55\% & 58\% & 50\% & 55\% & 59\% & 59\% & 38\% \\
  \hline
  \end{tabular}
  }
  \label{tab5}
  \vspace{-8pt}
  \end{table}

\begin{table}[t]
  \centering
  \begin{minipage}{0.48\textwidth}
    \centering
    \caption{Comparison with alternative method using RIS to provide pseudo segmentation labels for supervision.}
    \vspace{-10pt}
    \setlength{\tabcolsep}{0.5mm}
    \begin{tabular}{cc>{\columncolor{gray!10}}c}
    \specialrule{.5pt}{.1pt}{.1pt}
    ReferItGame & SegVG+RIS (LAVT) &  SegVG \\
    \specialrule{.1pt}{.1pt}{.1pt}
    val & 74.91  & 76.85 \\
    \specialrule{.1pt}{.1pt}{.1pt}
    test & 73.76 & 75.06 \\
    \specialrule{.1pt}{.1pt}{.1pt}
    \end{tabular}
    \label{tab6}
    \vspace{-12pt}
  \end{minipage}\hfill
  \begin{minipage}{0.48\textwidth}
    \centering
    \caption{
    AP50 using different segmentation queries for confidence score calculation on RefCOCOg-umd test set.
    }
    \vspace{-8pt}
    \setlength{\tabcolsep}{0.5mm}
    \begin{tabular}{c|c|c|c|c|c}
    \specialrule{.1pt}{.1pt}{.1pt}
    Seg Query & 1$^{\rm st}$ & 2$^{\rm nd}$ & 3$^{\rm rd}$ & 4$^{\rm th}$ & 5$^{\rm th}$  \\
    \specialrule{.1pt}{.1pt}{.1pt}
    AP50 & 84.65 & 84.57 & 84.73 & 84.78 & 84.63 \\
    \specialrule{.1pt}{.1pt}{.1pt}
    \end{tabular}
    \label{tab7}
    \vspace{-12pt}
  \end{minipage}
\end{table}

\subsection{Triple Alignment Analysis}
\vspace{-5pt}
In additional to the improvement results in ablation study, we further deepen our understanding of the Triple Alignment by analyzing the attention behavior. Specifically, we calculate the sum of the attention values from the query to the text-referred visual region (target bbox) as a percentage of the total attention (which includes attention to the query, text, and visual tokens) to illustrate the extent of alignment across these three modalities in the second, fourth, and final layer of Triple Alignment. We average the percentage across all the attention heads and queries, and perform the analysis across all the datasets. As shown in Table \ref{tab5}, in all the datasets, the attention values increase as the layer progresses, indicating that Triple Alignment progressively aligns the query to comprehend the text and then focus on the referred visual region. 

\subsection{Comparison with alternative method}
\vspace{-5pt}
Given the development of Referring Expression Segmentation (RES), a natural alternative method would be using a RES method to generate pseudo segmentation labels to substitute our bbox2seg paradigm. Therefore, to mimick real-world scenarios, we use LAVT~\cite{yang2022lavt} trained on RefCOCO to obtain pseudo segmentation labels on ReferItGame. We follow the same training setting in ablation study to conduct the comparison on ReferItGame. As shown in Table \ref{tab6}, our SegVG outperforms the alternative method. This demonstrates that our bbox2seg paradigm is more effective than using a RES model to provide pseudo segmentation labels which might suffer from the errors from the RES model.

\begin{figure}[t]
  \centering
  \begin{minipage}{0.5\textwidth}
      \centering
      \includegraphics[width=1\textwidth]{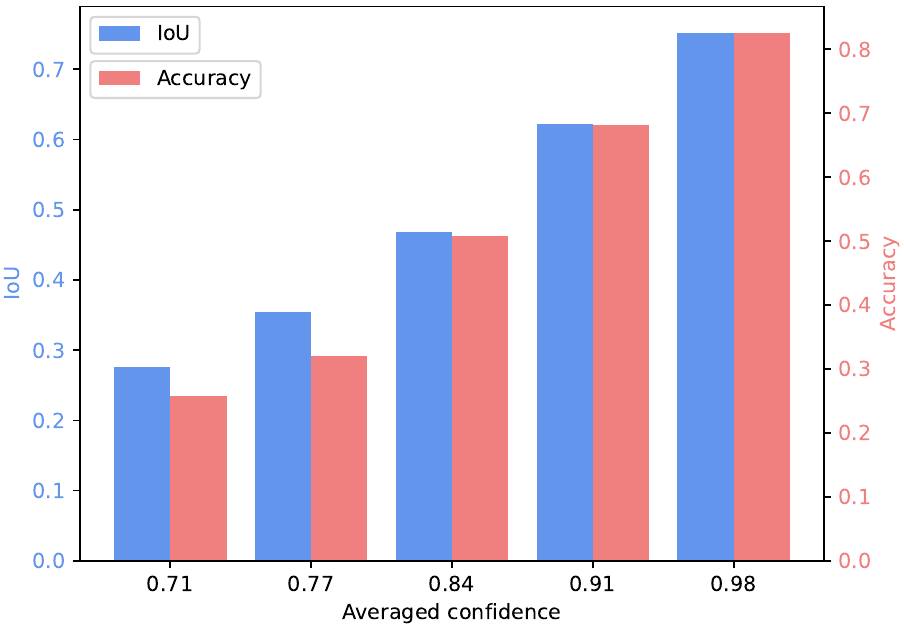} 
      \vspace{-16pt}
      \caption{IoU and Accuracy of different confidence scores on RefCOCOg-umd test set.}
      \label{conf_analyse}
      \vspace{-16pt}
  \end{minipage}\hfill
  \begin{minipage}{0.5\textwidth}
      \centering
      \captionof{table}{Performance of different confidence levels on RefCOCOg-umd test set.}
      \vspace{-2pt}
      \resizebox{1\textwidth}{!}{ 
      \setlength{\tabcolsep}{1mm} 
      \begin{tabular}{cccc}
          \toprule
          Confidence & IoU & Acc(\%) & Proportion (\%) \\
          \midrule
          $\geq 0.65$ & 0.7067 & 77.41\% & 100.00\% \\
          $\geq 0.70$ & 0.7072 & 77.48\% & 99.89\% \\
          $\geq 0.75$ & 0.7091 & 77.69\% & 99.44\% \\
          $\geq 0.80$ & 0.7143 & 78.37\% & 97.98\% \\
          $\geq 0.85$ & 0.7226 & 79.29\% & 95.04\% \\
          \bottomrule
      \end{tabular}}
      \label{reliability}
      \vspace{-16pt}
  \end{minipage}
\end{figure}

\subsection{Confidence Score Analysis}\label{conf_analyse_sec}
\vspace{-5pt}
In this section, we first detail the selection of segmentation query for the calculation of confidence score. Then, we evaluate the faithfulness of our confidence score. Finally, we demonstrate its utility in improving prediction reliability.

\paragraph{Selection of segmentation query}
To show the effect of different selection of segmentation query, we calculate AP50 on RefCOCOg-umd test set, using confidence score derived from each segmentation query. As shown in Table \ref{tab7}, the performance variations are slight among them. Therefore, we opt for the first segmentation query to calculate confidence score for simplicity. 
\vspace{-4pt}

\paragraph{Confidence Score Faithfulness}
To evaluate the faithfulness of our confidence score, i.e., whether a higher confidence score transformed from the segmentation output indeed indicates better performance, we assess the relationship of our confidence score and model performance metrics (IoU and Accuracy) as shown in Fig~\ref{conf_analyse}. We sort the RefCOCOg-umd test set by confidence score, split it into five equal parts, and calculate each part's average score and performance. We observe a positive correlation between the performance metrics and our confidence score, which confirms its faithfulness.
\vspace{-5pt}

\paragraph{Confidence Score Application}
In real applications, the confidence score can be used to enhance the model's reliability. Specifically, we can apply different confidence thresholds to achieve different predictions, as shown in Table~\ref{reliability}. First, we observe that accuracy increases with higher thresholds, indicating that adjusting the threshold can enhance the model's localization ability. Furthermore, the mean IoU also increases with the increasing threshold. Therefore, in downstream applications, such as using the model to provide pseudo-labels, we can increase the threshold to obtain a more accurate box. Since low-confidence outputs are excluded, the output proportion is slightly reduced, i.e., yields fewer outputs.
\vspace{-5pt}

\begin{figure}[t]
\begin{center}
    \centering
    \captionsetup{type=figure}
\includegraphics[width=0.86\textwidth]{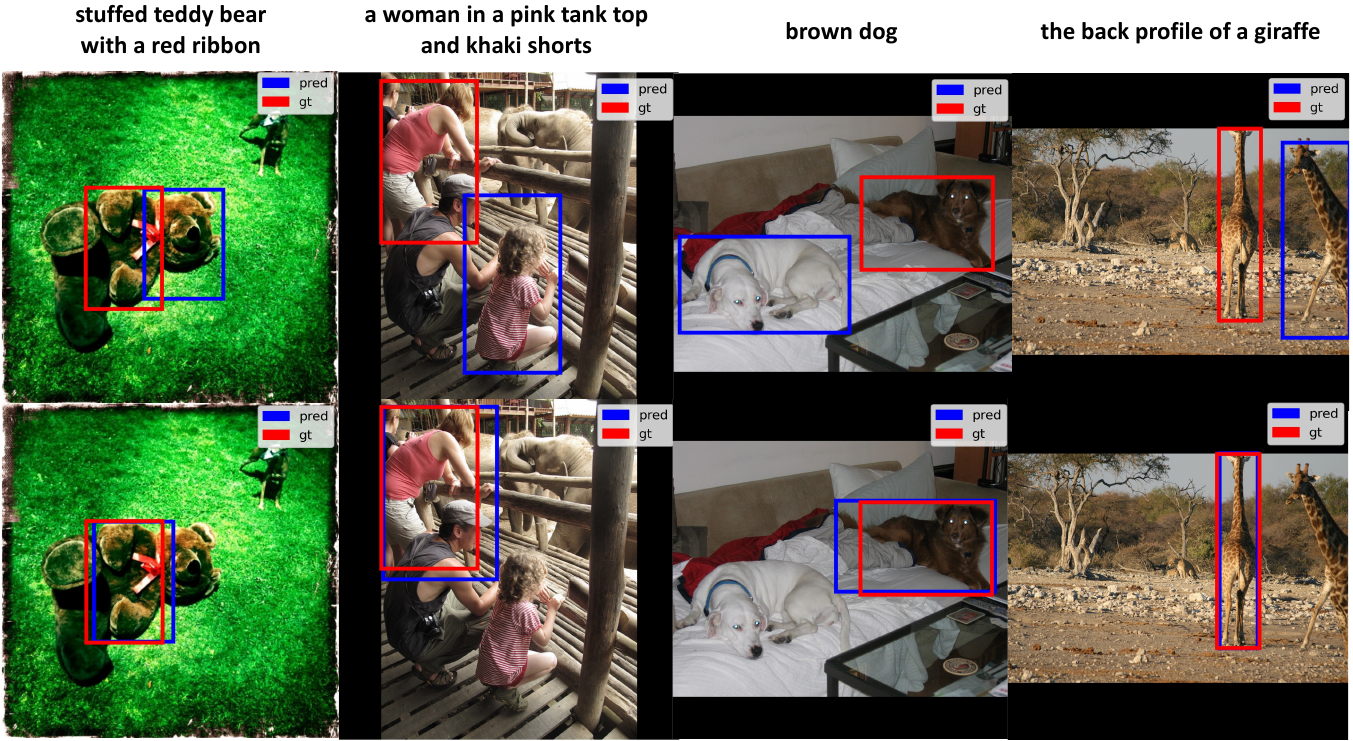}
\vspace{-10pt}
\caption{Qualitative comparison between {\bf (c)} (the first line) and {\bf (d)} (the second line) of Table.~\ref{ablation}. Red boxes are ground truth. Blue boxes are model predictions.}
\label{ablation_cd}
\end{center}
\vspace{-16pt}
\end{figure}

\begin{figure}[t]
\centering
\includegraphics[width=0.95\textwidth]{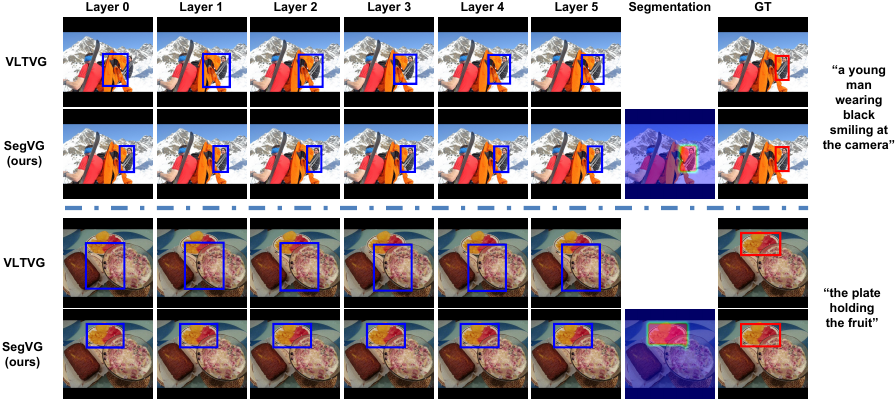}
\vspace{-10pt}
\caption{
Layer $i$ refers to the output of the $i$-th layer of the decoder. The blue boxes represent the models' predictions, while the red boxes denote the ground truth. In the segmentation mask shown in the second column from the right, red indicates high confidence for the foreground. Note that VLTVG does not provide segmentation output.
}
\label{Qualitative}
\vspace{-16pt}
\end{figure}

\subsection{Qualitative Results}\label{qualitative_res}
\vspace{-5pt}
Besides the ablation study's quantitative comparison showing that added segmentation queries improve robustness, we further compare (c) and (d) from Table~\ref{ablation} qualitatively to highlight the robustness enhancement. As shown in Fig.~\ref{ablation_cd}, adding segmentation queries improves the robustness of Decoder to distinguish the target from the distractor, such as the case of two ``dog'' or two ``giraffe''.

As depicted in Fig.~\ref{Qualitative}, we compare the box prediction quality by each decoding layer of SegVG with VLTVG~\cite{yang2022improving} which also involves multi-layer supervision. As seen in the upper two rows of Fig.~\ref{Qualitative}, VLTVG initially misses the target ``young man'' but improves its prediction gradually and finally makes the correct prediction. In contrast, due to the full exploitation of the annotations and the domain alignment in our Triple Alignment, SegVG successfully identifies the location of the target in the early decoding layer and consistently makes the correct prediction in each layer. Another example can be observed in the lower two rows of Fig.~\ref{Qualitative}. In this image, due to the complex colors, VLTVG fails to locate the target ``plate'' and consistently repeats the same mistake. Instead, SegVG correctly detects the target, even in the first decoder layer. Additionally, we visualize the segmentation mask obtained by SegVG in Fig.~\ref{Qualitative}, which accurately identifies the target box with high confidence. This behavior aligns with the box regression, demonstrating their shared objective, i.e., to distinguish the box.
\vspace{-10pt}
\section{Conclusion}
\vspace{-8pt}
We propose a new transformer-based model SegVG for Visual Grounding. Specifically, we introduce the Multi-layer Multi-task Encoder-Decoder to iteratively make the best of box annotations to involve pixel-level supervision. Moreover, we address the domain discrepancy among queries, text, and vision by the Triple Alignment module to improve subsequent target grounding. Extensive experiments demonstrate the superior performance of SegVG. Furthermore, we explore the reliability benefits of our segmentation output in real-world applications.


%
%
\bibliographystyle{splncs04}
\bibliography{main}
\end{document}